\documentclass[10pt,twocolumn,letterpaper]{article}

\usepackage[pagenumbers]{cvpr} 

\usepackage{graphicx}
\usepackage{amsmath}
\usepackage{amssymb}
\usepackage{booktabs}
\usepackage{subcaption}
\newcommand{\tablestyle}[2]{\setlength{\tabcolsep}{#1}\renewcommand{\arraystretch}{#2}\centering\footnotesize}
\newcommand{\improvement}[1]{\textcolor[rgb]{0.22,0.463,0.114}{\footnotesize{\textbf{#1}}}}
\usepackage[pagebackref,breaklinks,colorlinks]{hyperref}
\usepackage[ruled, linesnumbered]{algorithm2e}
\usepackage{colortbl}
\usepackage{balance}
\usepackage[capitalize]{cleveref}
\crefname{section}{Sec.}{Secs.}
\Crefname{section}{Section}{Sections}
\Crefname{table}{Table}{Tables}
\crefname{table}{Tab.}{Tabs.}

\def\cvprPaperID{622} 
\def\confName{CVPR}
\def\confYear{2023}

\begin{document}

\title{Query Efficient Cross-Dataset Transferable Black-Box Attack \\on Action Recognition}

\author{Rohit Gupta \textsuperscript{\rm 1} \\ 
{\tt\small rohitg@knights.ucf.edu}
\and Naveed Akhtar \textsuperscript{\rm 2} \\
{\tt\small naveed.akhtar@uwa.edu.au}
\and Gaurav Kumar Nayak \textsuperscript{\rm 3} \\
{\tt\small gauravnayak@iisc.ac.in}
\and Ajmal Mian \textsuperscript{\rm 2} \\
{\tt\small ajmal.mian@uwa.edu.au }
\and Mubarak Shah \textsuperscript{\rm 1} \\
{\tt\small shah@crcv.ucf.edu}
\and \textsuperscript{\rm 1}Center for Research in Computer Vision, University of Central Florida
\vspace{-1em}
\and \textsuperscript{\rm 2} University of Western Australia
\and \textsuperscript{\rm 3} Indian Institute of Science
}\maketitle

\begin{abstract}

    Black-box adversarial attacks present a realistic threat to action recognition systems. Existing black-box attacks follow either a query-based approach where an attack is optimized by querying the target model, or a transfer-based approach where attacks are generated using a substitute model. While these methods can achieve decent fooling rates, the former tends to be highly  query-inefficient while the latter assumes extensive knowledge of the black-box model's training data. In this paper, we propose a new attack on action recognition that addresses these shortcomings by generating perturbations to disrupt the features learned by a pre-trained substitute model to reduce the number of queries. By using a nearly disjoint dataset to train the substitute model, our method removes the requirement that the substitute model be trained using the same dataset as the target model, and leverages queries to the target model to retain the fooling rate benefits provided by query-based methods. This ultimately results in attacks which are more transferable than conventional black-box attacks. Through extensive experiments, we demonstrate highly query-efficient black-box attacks with the proposed framework. Our method achieves $8\%$ and $12\%$ higher deception rates compared to state-of-the-art query-based and transfer-based attacks, respectively. 
    
\end{abstract}

\section{Introduction}
\label{sec:intro}

\begin{figure*}[t]
  \centering
   \includegraphics[width=\linewidth, trim={0 0.3cm 0 0},clip]{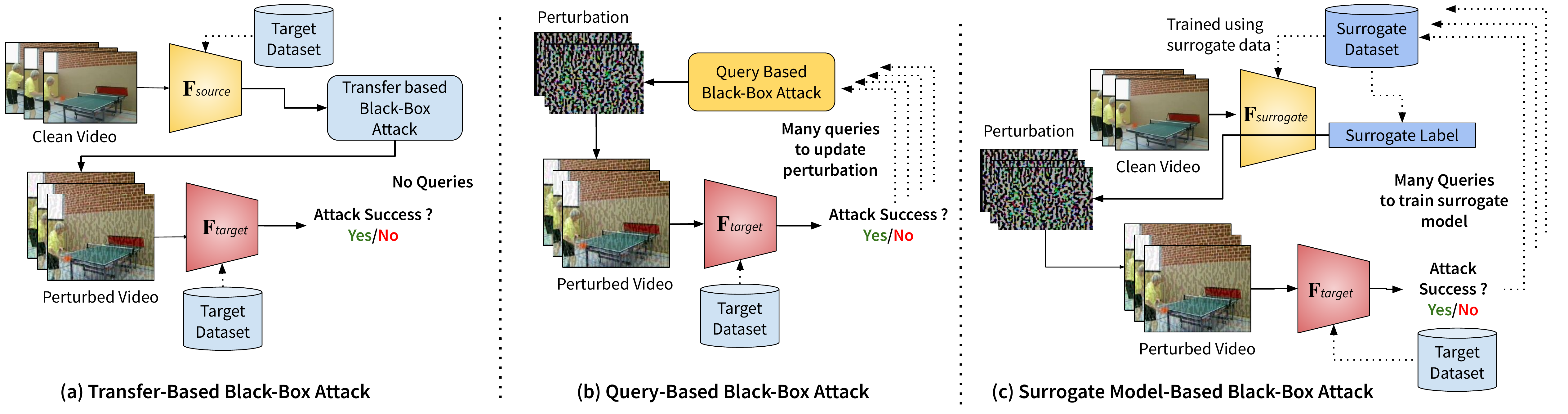}
   \vspace{-1mm}
   \caption{Existing approaches for generating black-box attacks: 
    \textbf{\textbf{(a) Transfer-based attacks} generate perturbations using a substitute model $F_{source}$ without querying the model. (b) Query-based attacks} generate a perturbation using many queries of the black-box model $F_{target}$. \textbf{(c) Surrogate model-based attacks} train a surrogate model to mimic the target black-box model, this requires many queries of the black-box to obtain the transfer set used for training the surrogate model. Query-based attacks typically achieve higher fooling rate than transfer-based attacks, however they often require tens of thousands of queries of the black-box (i.e. are not query-efficient).}
   \label{fig:conceptcomparison}
  \vspace{-4mm}
\end{figure*}

Deep convolutional~\cite{Hara_2018_CVPR, tran2019video, feichtenhofer2020x3d, xie2018rethinking} and Transformer networks~\cite{Arnab_2021_ICCV,Fan2021MVIT, gberta_2021_ICML, liu2022video, neimark2021video, li2022mvitv2} and hybrid architectures which combine them~\cite{xiao2021early} achieve high performance in recognizing actions from videos. Large-scale labeled datasets such as UCF101~\cite{soomro2012ucf101} and Kinetics~\cite{kay2017kinetics}, along with self-supervised learning methods like Spatio-Temporal Contrastive Learning~\cite{qian2021spatiotemporal} have played an important role in achieving these results. However, deep neural networks are vulnerable to adversarial attacks that can manipulate model predictions with imperceptible adversarial perturbations to inputs~\cite{goodfellow2014explaining}. Such attacks have also demonstrated success against action recognition models~\cite{wei2019sparse,cheng2018queryefficient,wei2020heuristic,Pony_2021_CVPR}. While typical ``white-box" adversarial attacks assume access to the target model weights, ``black-box" attacks~\cite{pmlr-v80-ilyas18a,brendel2018decisionbased,cheng2018queryefficient} do not require this access. 

There are three main approaches to black-box attacks, as illustrated in Fig.~\ref{fig:conceptcomparison}. The first assumes absolutely no access to the target model~\cite{papernot2017practical,inkawhich2019feature,wei2022towards,nowroozi2022demystifying,salman2020adversarially,cheng2019improving}. Methods in this category train a local substitute model then compute input perturbations to fool that substitute model. These computed perturbations are subsequently transferred to the unknown target model, hoping that they will retain their adversarial properties for the target model. This type of attack is commonly known as \textit{transfer-based} attack. The second stream assumes access to the outputs of the target model in the form of either the predicted label or the confidence scores of the output label~\cite{cheng2018queryefficient,pmlr-v80-ilyas18a,andriushchenko2020,dong2019,wei2020heuristic}.
In the former case, the black-box attack is called a \textit{hard-label attack}, whereas in the latter case it is called \textit{score-based attack}. Note that different naming conventions are often used for these attacks in the literature. For instance, the score-based attacks are referred to as partial-information attacks in \cite{pmlr-v80-ilyas18a} and hard-label attacks are called decision-based attacks in \cite{cheng2018queryefficient} and \cite{dong2019}. In this paper, we use the terms \textit{hard-label} and \textit{score-based} attacks. In both attack types, the attacker uses the information from the model's output to refine the adversarial examples. This requires making multiple queries to the target model to compute the adversarial input. Hence, both hard-label and score-based attacks belong to a common broader category of \textit{query-based} attacks.  

In practice, it is often the case that remotely deployed models (i.e.~target black-box models) do not provide full confidence score information to the users. Therefore, hard-label attacks are considered more pragmatic than the score-based attacks. Thus, we focus on the more challenging setting of hard-label attack in this work. A class of hybrid methods known as surrogate model-based attacks~\cite{zhou2020dast, sun2022exploring, lord2022attacking} have also been reported in the literature. These are transfer-based attacks which utilize a substitute model that is trained using surrogate data which has been distilled from the black-box model under attack. Although these models require a large number of queries to train the surrogate model, subsequent attacks can be generated without any queries.

The goal of an ideal black-box attack is to fool the deployed models without any information about them. However, cuerent transfer- and query-based  attacks (and their hybrids) fall short on truly satisfying the requirements of a perfect black-box setting. A vast majority of the contemporary transfer-based attacks only consider transfer between different models of the same data distribution. Most commonly, the transfer is evaluated across different models trained on ImageNet~\cite{Su2018,inkawhich2019feature}. This results in two major assumptions to be made about the target model. First, the attacker knows the training data samples, and second, the attacker is even aware of the ontology of class labels of the target model. This weakens the strict `black-box' setup. However, these violations have not been addressed by prior methods. We address these limitations in formulating our black-box setting.

For the query-based attacks, contemporary methods do not impose an upper bound on the number of queries and it is common to allow even ten thousand queries~\cite{Chen2017Zoo,wei2020heuristic}. Theoretically, a sufficiently large number of queries can estimate the training data distribution of the target model~\cite{sanyal2022towards, zhang2022towards, truong2021data}. Hence, the absence of upper bound on queries is not only in violation of strict `black-box' setting, but is also inappropriate for an attack type that is generally marketed as a `practical' threat. This problem is particularly acute for video classification, where inference is much more expensive than that for images because of higher data  dimensionality. Moreover, despite extensive research on image-based attacks, adversarial attacks on action recognition systems are still under-explored.  

In this work, we address the above-noted issues with a novel framework for fooling action recognition models under minimal (practical) assumptions about the target model, and extend that to a query-efficient hard-label attack. The proposed \textit{Video Representation Attack} framework - illustrated in Fig.~\ref{fig:system} - not only removes the impractical assumptions from the transfer-based attacks regime, but also exhibits a seamless extension to the query-based setup. From the transfer-based attack perspective, we utilize a source model trained using a dataset with limited overlap with the target model. Perturbation for videos are computed to disrupt the feature space of this source model. To extend the attack to the query-based scenario, we employ a Gram-Schmidt Orthogonalization scheme that chooses the direction of perturbations to fool the target model by querying it. Our method require less than a dozen queries achieve fooling rates similar to prior query based methods with thousands of queries.
The key contributions of our work are:
\vspace{-2mm}
\begin{itemize}
\setlength\itemsep{-0.1em}
\item We are the first to study cross-dataset transfer of black-box attacks for action recognition models.
\item We propose crafting adversarial samples to disrupt the feature representation of videos to improve cross-dataset transferability of the attack  (Sec.~\ref{sec:method}).
\item Utilizing iterative Gram-Schmidt Orthogonalization, we extend our attack to an efficient query-based attack. 
\item We establish the efficacy of our attack on four benchmark datasets (Kinetics400, Something-Somethingv2, UCF101 and HMDB51) and 6 architectures (4 CNN \& 2 Transformers) (Sec.~\ref{sec:results}). Our attack achieves a 7\% higher fooling rate on Kinetics than transfer-based attacks while using only 10 queries.
\end{itemize}

\section{Related Work}
\label{sec:related}
\vspace{-2mm}
We focus on pioneering and works closely related to ours in this related works section. Adversarial attacks and defenses are broad areas actively being investigated. For a more thorough outline of the recent advances in the adversarial domain, we refer interested readers to~\cite{akhtar2021advances}. 

\subsection{Black-Box Attacks on Image Classification}

Prior work on black-box attacks can be divided into two major categories: \textit{transfer-based} and \textit{query-based} attacks. The latter can further be sub-categorised into \textit{score-based} and \textit{hard-label} attacks.

\vspace{1mm}
 \noindent{\bf Transfer-based Attacks:} Work on transfer-based attacks is typically focused on investigating transferability of adversarial examples across architectures trained on the same dataset~\cite{papernot2017practical, Zhao2021Success}, or generalizations of this setting where alternative data is collected and used based on the knowledge of the class ontology of the classifier. In this cross-architecture, same-dataset setting, it has been found that momentum boosting (\textbf{MI-FGSM})~\cite{dong2018boosting} and use of augmentations during attack iterations (\textbf{DI-FGSM})~\cite{xie2019improving} lead to a more transferable attack. The Activation Attack~\cite{inkawhich2019feature} method which aims to transform the internal activations of the model for an image to be similar to the activations of the model for a different image from the target class is reported to transfer better across architectures. This notion is conceptually related to our technique. However, \cite{inkawhich2019feature} does not deal with video data or cross-dataset transfer of attacks. It also requires access to samples from the target class in order to generate attacks.

\vspace{1mm}
 \noindent{\bf Query-based Attacks:} Classifier score-based attacks~\cite{Chen2017Zoo} assume access to class confidence scores predicted by the model and rely on estimating gradients of the class confidence scores with respect to the input. Estimating these gradients require a large number of queries (roughly of the order of $10,000$ queries per image) of the black-box model.  Successive works~\cite{ilyas18a} have generally focused on improving the query efficiency of gradient estimation. Boundary attack~\cite{brendel2018decisionbased} was the first general technique for computing  imperceptible hard-label black-box attacks. It performs a random walk starting from a random perturbation that already causes a misclassification, significantly reducing its perceptibility in the process. However, this process also requires a very large number of queries (sometimes $>100,000$) to attack a single image. The somewhat more query efficient OPT-attack~\cite{cheng2018queryefficient} formulates a proxy optimization problem in order to generate the black-box adversarial attack. The inefficiency of query-based attacks is particularly acute in the case of video classification models, due to the higher dimensionality and temporal redundancy in videos.

\subsection{Adversarial Attacks on Action Recognition} Adversarial attacks, especially black-box adversarial attacks on action recognition models are significantly harder than attacks on image tasks since there is an additional dimension and temporal redundancy. 

\noindent \textbf{White-box Attacks:} Wei \etal.~\cite{wei2019sparse} investigated white-box adversarial attacks on video classification models which produce sparse perturbations. Li \etal.~\cite{li2019stealthy} developed a white-box attack for realtime streaming video classification systems which can fool a classifier despite being stochastically temporally shifted or misaligned.

\noindent \textbf{Transfer-based Attacks:} Flickering attacks~\cite{Pony_2021_CVPR} can be physically implemented through control of scene illumination and are also tested in a transfer-based setting.

\noindent \textbf{Score-Based Attacks:} Wei \etal.~\cite{wei2020heuristic} designed a score-based black-box attack that utilizes heuristics to select a subset of video frames to attack in order to reduce the number of queries required (henceforth referred to as OPT+Heuristics). They also provided results for an extension of the OPT-attack~\cite{cheng2018queryefficient} to videos, which requires twice as many queries. Motion priors from optical flow~\cite{zhang2020motion} have also been used to improve the efficiency of attacks. Geo-TRAP~\cite{Li2021Adversarial} uses geometric transformation operations to restrict the search space of video adversarial perturbations and improve the query efficiency of score-based attacks.


Whereas plenty of attack methods are available to fool image models, attacks on video action recognition, especially in the \textit{hard-label black-box setting}, are still under-explored. The added temporal dimension in videos not only makes this problem computationally more demanding, but also makes achieving the fooling objective harder. Moreover, due to their inspirational roots in image-based attacks, black-box attacks in action recognition also suffer from inappropriate assumptions, such as, restricted domain shift and excessive number of queries, that violate practical black-box setups. We address these issues in this work.





\vspace{-2mm}
\section{Method}
\label{sec:method}
\vspace{-2mm}

\begin{figure*}
\centering
\begin{minipage}{.59\textwidth}
  \centering
  \captionsetup{width=.9\linewidth}
  \includegraphics[width=\linewidth, trim={0 0.3cm 0 0},clip]{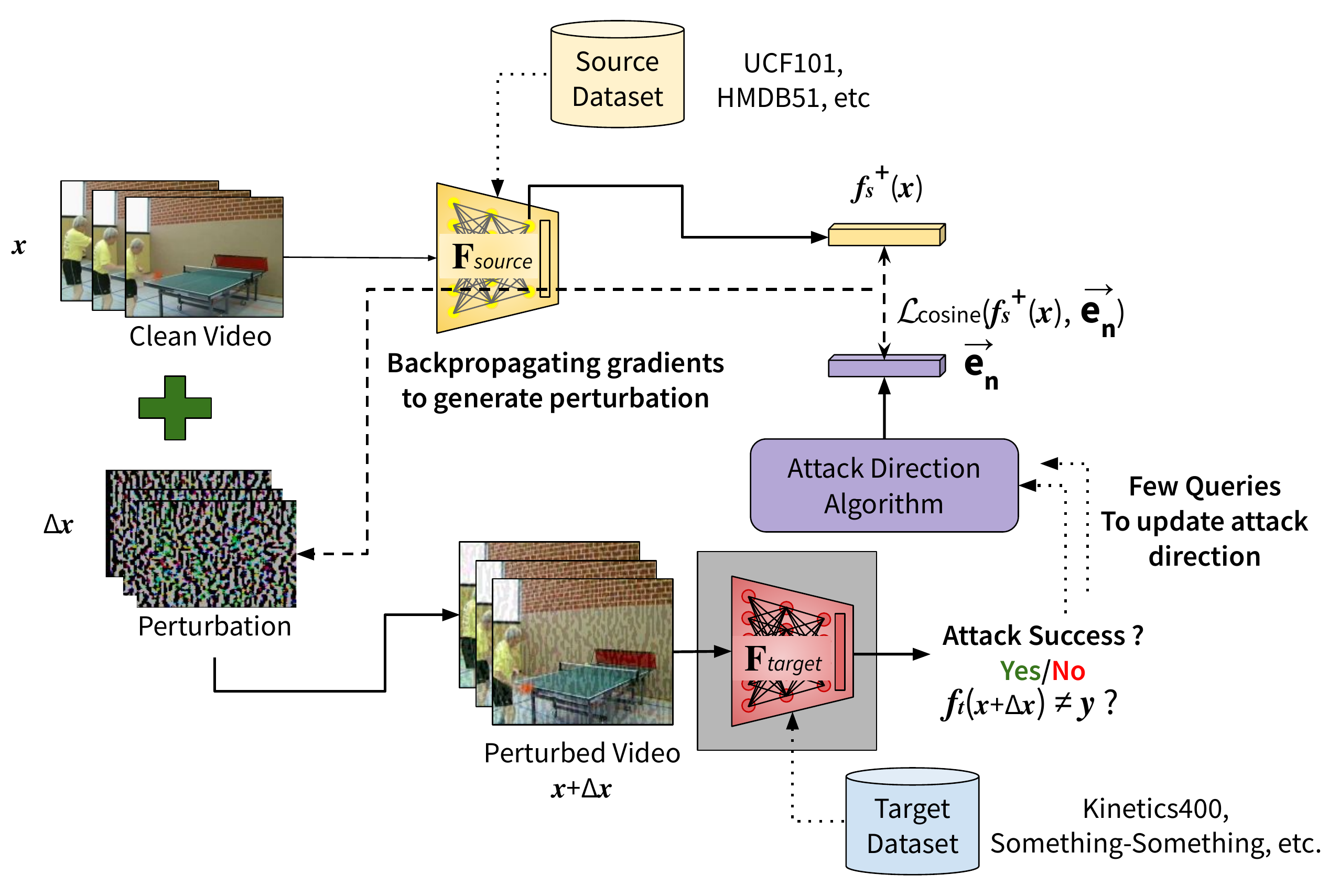}
   \vspace{-7mm}
   \caption{\textbf{Video Representation Attack (VRA)} to unify the benefits of query-based and transfer-based approaches to achieve query efficient attack in a hard cross-dataset attack setting. VRA targets the internal representation of source model ($\mathit{f_s}$) in order to generate transferable perturbations ($\Delta x_i$). The attack direction ($\vec{e_i}$) in which the representation is to be shifted is orthogonalized using  Gram-Schmidt process. Cosine loss between the original representation and the attack direction is used as the objective, and the perturbation is generated by backpropagating the gradients to the input. The perturbed video is then sent to the black-box target model ($\mathit{f_t}$) as a query to verify if the attack succeeded. The attack can be iterated as a multi-query attack by varying the attack direction to find a perturbation to successfully fool the black-box.}
  \label{fig:system}
\end{minipage}
\begin{minipage}{.4\textwidth}
  \centering
  \vspace{3mm}
  \captionsetup{width=.9\linewidth}
  \includegraphics[width=\linewidth]{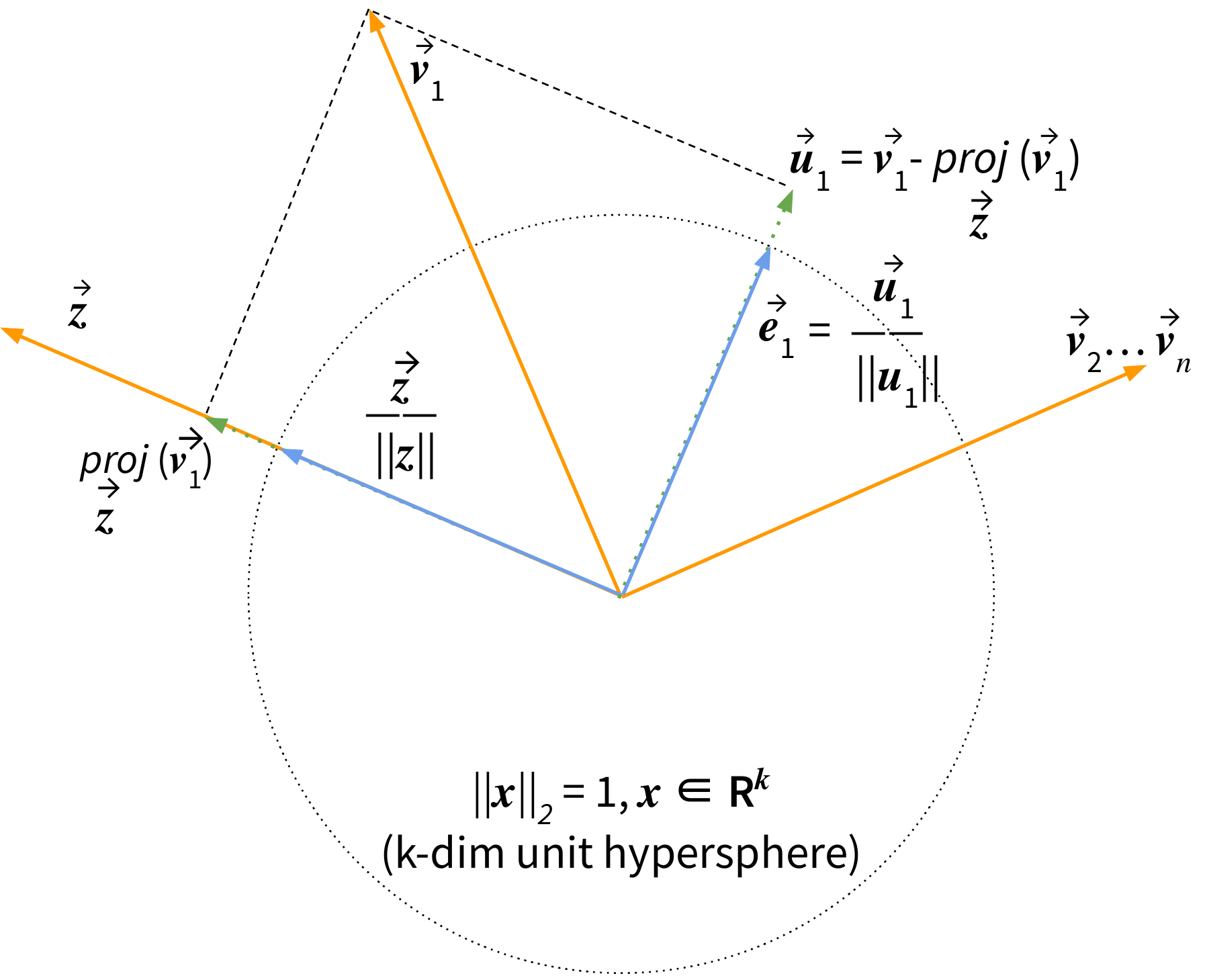}
  \vspace{5mm}
   \caption{One step of \textbf{Orthogonal Attack Direction (OAD) search}. $\vec{z}$ is the feature representation of the video from penultimate layer being attacked, whereas $\vec{v_1}, \vec{v_2}, ... ,\vec{v_n}$ are sampled from a psuedo-random number generator. $\vec{u_1}$, the component of $\vec{v_1}$ orthogonal to $\vec{z}$, is obtained via projection, and is normalized to generate the first attack direction $\vec{e_1}$. Attack directions $\vec{e_1},  ... ,\vec{e_n}$, are mutually orthogonal and lie on the same hypersphere as the video representations.}
  \label{fig:OADS}
\end{minipage}
\vspace{-5mm}
\end{figure*}

We first introduce the cross-dataset setting for our adversarial attack on action recognition models. We carry out attacks using a \textit{source} model trained on a different dataset than the \textit{target} model while assuming only limited label overlap between the datasets. Next, we introduce our proposed Video Representation Attack (Fig.~\ref{fig:system}) to generate transferable perturbations under this black-box setting. The attack is guided by chosen perturbation directions. We eventually develop a Gram-Schmidt Orthogonalization based algorithm for adjusting the perturbation direction, which allows enhancement of our method to a query efficient attack for action recognition.

\vspace{-1mm}
\subsection{Problem Formulation}
\label{sec:formulation}
\vspace{-2mm}

In  black-box deployment of action recognition system, the user can supply any input RGB video `$x$' to a black-box classifier, $f_t(x): \mathbb{R}^{T\times H \times W \times 3} \rightarrow \mathcal{Y}$, and receive the predicted most likely class label $\hat{y}$ from the set of class labels $\mathcal{Y}$. In our settings, there  is no knowledge available regarding the predicted class probabilities $\mathcal{P}_t(y|x)$, or the model gradients with respect to input $\nabla_x \mathcal{P}_t(y|x)$, which is often employed to compute adversarial inputs. 
Considering practical scenarios, we also put  a limit on the number of queries $Q_{max}$ that can be made to the classifier.  

\vspace{2mm}

{\noindent \textbf{Transfer-based Attacks}:} In the standard transfer-based black-box setting~\cite{papernot2017practical}, the set of class labels of the black-box, $\mathcal{Y} = \{y_1, ... , y_n\}$, is known to the attacker. A substitute or source classifier, $f_s(x): \mathbb{R}^{T\times H \times W \times 3} \rightarrow \mathcal{Y}$ is trained on the dataset used to train the target black-box model or an alternate dataset with the same class labels. The transferable perturbation is then generated using this source model. {\em In contrast, in our setting, the source model $f_s(x)$ is trained on a dataset  that is completely different from the target model training data, and also has a different set of class labels: $ \mathcal{S} = \{s_1, ... , s_k\}$}. In this case, we allow partial overlap between the two sets of labels ~$| \mathcal{S} \cap  \mathcal{Y}| \leq |\mathcal Y| \land (\mathcal S \cap \mathcal Y) \neq \emptyset $. 
This corresponds to the practical situation where the attackers may have knowledge of a subset of classes of the target model but do not have an exhaustive list of the  classes recognized by the black-box classifier. In this setup, where the attacker has knowledge of as few as one label, the task of launching an untargeted  adversarial attack can be modelled as computing a perturbation $\Delta x$ which minimizes the target model's predicted probability of the correct video label by the target model:
\begin{equation}
    \underset{\Delta x}{\mathrm{\min}}~\mathcal{P}_t(y\mid(x+\Delta x)),~\text{s.t.}~||\Delta x||_p < \epsilon,
\end{equation}
where $||.||_p$ denotes the $\ell_p$-norm of the vector and $\epsilon$ is a pre-defined scalar. We let $p = \infty$ in this work because other norm bounds are known to result in the differences of limited practical value~\cite{kurakin2016adversarial, moosavi2016deepfool}.  
In the standard transfer-based attacks, the class labels for the source and target models are the same, hence the task simply changes to
\begin{equation}
    \underset{\Delta x}{\mathrm{\min}}~\mathcal{P}_s(s\mid(x+\Delta x)),~\text{s.t.}~||\Delta x||_\infty < \epsilon,
    \label{eq:min2}
\end{equation}
where $\mathcal P_s(.)$ indicates the predicted probability of the source model. In our setting, the label space $\mathcal S$ of the source model does not perfectly match $\mathcal Y$.
Considering $\mathcal P_{s/t}(.)$ under the Kolmogorov axioms of probability, our objective is harder than the traditional transfer-based attacks.
Since $\forall i, \sum_i\mathcal P_t(y_i|x) = 1$ and $\forall j, \sum_j\mathcal P(s_j|x) = 1$, but $\mathcal S \neq \mathcal Y$, thus minimization of  $\mathcal{P}_s(s|(x+\Delta x))$ in Eq.~\eqref{eq:min2} does not necessarily imply fooling of the target model. This calls for disrupting the internal features of `$x$' for the fooling purpose to ensure that the target model is not able to make the correct prediction. 

Under the proposed setup, a more desirable option is to have some feedback from the target model about its predicted label, to ensure that we are indeed able to minimize $\mathcal P_t(y|x)$. This leads to the natural extension of our objective to the query-based attack.  

\vspace{2mm}
{\noindent \textbf{Query-based Attacks}:} Query-based attacks, such as the Boundary Attack~\cite{brendel2018decisionbased} and NES Attack~\cite{ilyas18a}, typically begin with a noise signal or the target class sample which is already ``\textit{misclassified}" by the target model, and iteratively refine it to match the input used to  attack the model. No assumptions are made about the target model. Moreover, typical query-based attack algorithms also do not attempt to train a substitute model. Most of the query-based methods, in order to be efficient, rely on a psuedo-gradient estimation of the target model's unknown loss function with respect to the input pixels. Since the dimensionality of the input $x \in \mathbb{R}^{T\times H \times W \times 3}$ is very high in the case of videos, estimating psuedo-gradients naturally requires a much larger number of queries. This makes the typical query-based attack schemes impractical for this setting, as they lead to computational intractability.

The key idea we adopt from the query-based attacks is the notion of exploring the black-box model's properties through  varying query directions obtained by random sampling. However, instead of utilizing random queries in the input space, we utilize our substitute model to generate query perturbations by manipulating the representation of the input. This substitute model is available to us because we induce this as the source model for our transfer-based attack (without querying the target model). In this case, each perturbation corresponds to a shift of the input along a certain direction in the feature space of the substitute model. As per the manifold assumption under-pining most deep representation learning methods, directions in the feature space of a well-trained model typically represents a semantic shift in the input data. Another perspective on this perturbation is that different attack directions represent different selections of the representation features to attack. Since a transfer-based attack relies on the fact that the source and target models have some common feature representation, exploring multiple attack directions in the representation space should allow discovery of the  relevant features to distort. 







\vspace{-1mm}
\subsection{Proposed Transfer-based Attack }
\label{sec:attack}
\vspace{-2mm}

A key strength of transfer-based attacks is that they are able to utilize prior knowledge about the data domain in the form of learned substitute or source model(s). On the other hand, the quintessential property  of query-based attacks is that they are iterative and are able to incorporate feedback of their queries to the target model, generally as some form of psuedo-gradient estimation. Query-based attacks become monotonically stronger with an increased number of queries. This feature is generally considered incompatible to transfer-based attack paradigm. However, in our opinion, these two key strengths of the two paradigms are complementary. Thus, we conjecture that an attack that is able to combine these two strengths would be a pareto-improvement  over the status quo. This is the key concept behind our Video Representation Attack (VRA) framework.

In the proposed attack, to compute transfer-based adversarial examples, we leverage the intuition on dissimilarity between our objective and the objective of standard transfer-based attacks  discussed in Sec.~\ref{sec:formulation}. That is,  instead of generating the perturbation to flip the label predicted for the video by the source model, we generate the perturbation such that the feature space representation of the video shifts in a specified adversarial direction. This  direction can be a vector chosen randomly from a uniform distribution with the same dimensions as the representation space of the source model (i.e. $\vec{\mathbf{v}} \sim \mathcal U[0, 1]$). The feature vector, $f_s^+(x)$ and  attack direction, $\vec{\mathbf{v}}$ are $L_2$ normalized, allowing the use of cosine loss to measure the distance between the vectors as: \begin{equation}
    \mathcal{L}_{VRA}(f_s^+(x), \vec{\mathbf{v}}) =  \frac{f_s^+(x) \cdot \vec{\mathbf{v}}}{\|f_s^+(x)\| \|\vec{\mathbf{v}}\|}.
    \label{eq:dfdrand}
\end{equation}

We utilize the above-noted loss to estimate a gradient-based perturbation as follows
\begin{equation}
    \Delta x = - \epsilon \cdot sign(\nabla_x \mathcal{L}_{VRA}(f_s^+(x), \vec{\mathbf{v}})).
    \label{eq:gradstep}
\end{equation}
Notice in the above equation that we utilize model gradients. Those are available to us because currently we deal with the source model about which we have complete information. What makes our source model different from the conventional source models is the fact that we are not making any assumptions about knowledge of the training data or the label ontology of this model. In our experiments (Sec.~\ref{sec:results}), we demonstrate that our attack achieves excellent performance under this real-life setting.

\vspace{3mm}
\noindent \textbf{Single Step Attack:} Our attack uses only a \textit{single gradient step} to generate the perturbation. Prior work on transfer-based attack in the same dataset, cross-architecture transfer setting has found that when using input augmentations~\cite{xie2019improving}, momentum boosting~\cite{dong2018boosting} and other methods to prevent overfitting, multi-iterated attacks can work better than single step attacks. We find empirical evidence to demonstrate that using multiple iterations leads to poor transferability in the cross-dataset transfer setting. This applies not just to our attack but also to label based attacks.

\vspace{2mm}
\noindent \textbf{Multi-Temporal, Multi-Layer Attack:} Our attack can also be carried out with features taken across multiple time steps and from across multiple layers of the source model. These features can be concatenated into a multi-temporal, multi-layer representation and attack can proceed as usual. \\

$f_{multi} = [f_s^+(t_1, l_1),  f_s^+(t_2, l_1),  f_s^+(t_1, l_2),  f_s^+(t_2, l_2)].$

\subsection{Extension to Query-based Attack}
\vspace{-2mm}
Although our attack may at first seem a variation of conventional transferable attack, it is carefully designed to hold a key property for its extension to query-based attack. Unlike typical untargeted label-flipping attacks, our attack has the potential to explore multiple attack directions in the feature space of our source model. That is, in Eq.~\eqref{eq:gradstep}, we can control $\vec{\mathbf{v}}$. This is ideal for incorporating feedback in $\Delta x$, provided we have the means to get that feedback from the target model. The query-based attack paradigm allows exactly getting that feedback. As we operate in the challenging and pragmatic hard-label attack scenario in this work, only the final predicted label from the black-box is used for feedback. Extension of our transfer-based attack to query-based attack boils down to a systematic iterative search of $\vec{\bf v}$, that monotonically reduces $\mathcal{P}_t(y|x)$ by incorporating the predicted label of the target model. We perform this search with an iterative algorithm that is discussed next.

\vspace{-2mm}
\subsection{Optimizing the Attack Direction}
\label{sec:orthogonal}
\vspace{-2mm}

Our framework facilitates generation of multiple perturbations for a given input by varying $\vec{\bf v}$ in Eq.~\eqref{eq:pertrand}. The success of these perturbations can be verified by querying the target model. We conceptualize this  querying process as a search process with an objective to find a successful adversarial direction, given a target model and input. However, because of the high dimensionality of the representation space, a purely random search process is inefficient. Moreover, random search discards the information gained from the failed previous searched directions. In order to address that, we propose an algorithm to systematically search for our desired vector by iteratively exploring orthognoal directions. We refer to it as the Orthogonal Attack Directions (OAD) algorithm. 

The OAD improves on simple random search by ensuring that each successive attack direction is orthogonal to all the prior attack directions, thereby building on the earlier queries. Gram-Schmidt method~\cite{Gram1879} is used to ensure this orthogonalization  in the feature space of the source model.

\RestyleAlgo{ruled}
\SetKwComment{Comment}{/* }{ */}
\setlength{\textfloatsep}{3pt}
\begin{algorithm}[t]
\caption{Our attack on target model $f_t$ in query-based setup, using source model $f_s$}\label{alg:dfd}
\KwData{$Q_{max} \geq 1$, $x \in \mathbb{R}^{T\times H\times W\times 3}$}
\KwResult{$\Delta x \ni f_t(x + \Delta x) \neq y $}
$i \gets 1$\;
$\vec{v_0} \gets f_s^+(x)$\;
\While{$i \leq Q_{max}$}{
   $\vec{\mathbf{v}_i} \sim U[0, 1]$\;
    $\vec{\mathbf{u}_i} = \vec{v_i} - \sum_{j=1}^{i-1}\mathrm{proj}_{\vec{u_j}} (\vec{v_i})$\;
    $\vec{\mathbf{e}_i} = \frac{\vec{u_i}}{\|\vec{u_i}\|}$\;
  $\mathcal{L}_{VRA} =  \frac{f_s^+(x) \cdot \vec{\mathbf{e_i}}}{\|f_s^+(x)\| \|\vec{\mathbf{e_i}}\|}$\;
  $\Delta \mathit{x} = - \epsilon \cdot sign(\nabla_x \mathcal{L}_{VRA}(f_s^+(x), \vec{\mathbf{e_i}}))$\;
  \If{$f_t(x + \Delta x) \neq y$}{
    \Return success\;
  }
  $i \gets i + 1$\;
}
\end{algorithm}

We provide a simplified visual illustration of our technique in Fig.~\ref{fig:OADS}. The  process  of computing the direction (following the figure) is as below:
\begin{enumerate}
\setlength\itemsep{-0.01em}
  \item Initialize $\vec{v_0} = f_s^+(x)$ as the representation of the video in the feature space of the source model. 
  \item Sample $\vec{v_1}, \vec{v_2},...,\vec{v_n}$  from a uniform distribution $\mathcal U[0,1]$ with a psuedo-random number generator. 
  \item These vectors are not guaranteed to be orthogonal and hence need to be orthogonalized. 
  This is achieved by projecting the current raw random vector $\vec{v_n}$ along all the previous vectors $\vec{v_i}, i \in 1 ... n-1$ and selecting only the component $\vec{u_n}$ of $\vec{v_n}$ which is orthogonal to all prior $\vec{v_i}, i \in 1 ... n-1$ as the next attack direction. Formally, 
 \vspace{-3mm}
  \begin{equation} \label{eq:gramschmidt}
\vec{\mathbf{u}_n} = \vec{\mathbf{v}_n} - \sum_{j=1}^{n-1}\mathrm{proj}_{\vec{\mathbf{u}_j}} (\vec{\mathbf{v}_n}).
\end{equation}
\vspace{-4mm}
  \item Since the feature vectors are constrained to lie on the unit-norm hypersphere, the normalized vectors $\vec{e_n}$ are obtained by appropriately scaling $\vec{u_n}$. Now, vectors $\vec{e_1}$, $\vec{e_2}$ ... are the actual attack directions which can be used. 
  \item The steps of the attack direction generation are repeated until a successful perturbation is found, or the number of queries exceed pre-specified limit $Q_{max}$.
\end{enumerate}

\noindent In the above process, each  attack direction represents a different combination of the source representation features to modify. For the success of adversarial attack, we need to discover common features that are salient to both source and target models for the action classification task.
Hence, sing multiple search steps strengthens the attack. The use of orthogonal attack directions helps in quickly exploring the search space, thereby reducing the number of queries. 

We summarize the process of launching our attack in query-based setup as Algorithm~\ref{alg:dfd}. Since 3D-CNN action recognition models are used for generating the video perturbation  $\Delta x \in \mathbb{R}^{T\times H\times W\times 3}$, it has a spatio-temporal structure which disrupts both motion and  appearance characteristics of the input. The main constraint we impose on the perturbation is the hard cap on $L_\infty$ norm, but if required, other constraints from the literature such as sparsity~\cite{wei2019sparse} and temporal smoothness~\cite{Pony_2021_CVPR} can also be incorporated. 


\vspace{-2mm}
\section{Evaluation}
\label{sec:results}
\vspace{-2mm}

\noindent{\textbf{black-box Target Models:}} In order to create a realistic black-box setting, we need models that are not trained locally. Hence, we utilize Kinetics400~\cite{kay2017kinetics}  and Something-Somethingv2~\cite{goyal2017something} pre-Trained models from the TorchVision~\footnote{https://pytorch.org/vision/0.14/models.html\#video-classification}~\cite{paszke2019pytorch} and PyTorch Video~\footnote{https://pytorchvideo.org/} model zoos. 

Six different architectures are available: 3D-ResNet (R3D)~\cite{Hara_2018_CVPR}, ResNet-(2+1)-D (R(2+1)D)~\cite{tran2019video}, Mixed Convolution Net (MC3)~\cite{tran2019video}, Separable 3D-CNN (S3D)~\cite{xie2018rethinking},  Multiscale Vision Transformers (MViT) v1~\cite{fan2021multiscale} and v2~\cite{li2022mvitv2}. These models are widely used for different video understanding tasks, and hence are a valuable targets for any adversarial attack.

\vspace{0.5mm}
\noindent{\textbf{Source Models:}} For generating our attacks, we utilize substitute (i.e.~source) models trained on UCF101~\cite{soomro2012ucf101} and HMDB51~\cite{Kuehne2011hmdb} action recognition datasets. Exact or near-exact matches for 70 out of 101 UCF101 classes are present in Kinetics400, while 29 out of 51 HMDB51 classes match is present in Kinetics400. This makes them a perfect fit as a test case for our limited label overlap setting.

\vspace{0.5mm}
{\noindent \textbf{Evaluation Metrics}:} We use the Attack Success Rate (\textbf{ASR}) and Deception Rate (\textbf{DR}) of the target model to evaluate the effectiveness of the attacks. The deception rate corresponds to the fraction of otherwise correctly identified videos for which the attack successfully fools the target model. Formally,  
 \begin{equation} \label{eq:deceptionr}
\mathbf{DR} = \frac{\hat{P}(y|x) - \hat{P}(y|x+\Delta x)}{\hat{P}(y|x)},
\end{equation}

 \begin{equation} \label{eq:asr}
\mathbf{ASR} = 1 - \hat{P}(y|x+\Delta x),
\end{equation}
where $x$ and $y$ correspond to the clean video input and label respectively, and $\Delta x$ is the perturbation generated by the attack. $\hat{P}(y|x)$ represents the Top-1 accuracy of making the correct prediction averaged over the validation set.

\vspace{0.5mm}
{\noindent \textbf{Baselines}:} Prior work in cross-architecture transferability in the image domain has found that multi-iterated attacks (I-FGSM)~\cite{kurakin2016adversarial} overfit to the source model and have lower transferability. Attacking least likely predicted class (LL-FGSM)~\cite{kurakin2017adversarial}, using momentum boosting (MI-FGSM)~\cite{dong2018boosting} and diverse augmented inputs (DI\textsuperscript{2}-FGSM)~\cite{xie2019improving} helps to prevent overfitting and boosts iterated attacks to have higher transferability. We test all these baselines and their combinations, along with the video-specific Flickering Attack~\cite{Pony_2021_CVPR}. As we find that the single-iteration LL-FGSM baseline is the strongest baseline attack, we also develop a query-based variant of the Targeted-LL-FGSM attack which successively tries to perturb the prediction to each of the classes of the source model. The number of queries this baseline can generate is of course limited by the number of classes. e.g. for UCF101 models, this baseline can only generate 100 different perturbations per sample. We also test a random perturbation baseline for reference. 
Note that as our attacks are in the \textit{hard-label} setting, we do not compare to \textit{score-based} attacks like OPT~\cite{cheng2018queryefficient, wei2020heuristic}, Motion-Excited Sampler~\cite{zhang2020motion} and Geo-TRAP\cite{Li2021Adversarial} which require access to predicted class probabilities from the black-box.

\begin{table}[t]
\tablestyle{2pt}{0.95}
\centering
\small
\begin{tabular}{lccc} 
\toprule
\textbf{Method} & \textbf{Queries} & \textbf{ASR}   & \textbf{DR}   \\ 
\toprule
\multicolumn{4}{c}{\textbf{Transfer-based Methods}} \\ 
\cmidrule{1-4}
Flickering\textbf{ }(\textbf{$\epsilon$ = 0.2})~\cite{Pony_2021_CVPR} & 1 & 49.0 & 15.20 \\ 
 FGSM & 1 & 49.3 & 15.65 \\ 
 I-FGSM & 1 & 46.3 & 10.54 \\ 
 MI-FGSM & 1 & 47.8 & 13.09 \\ 
 DI$^2$-FGSM & 1 & 47.4 & 12.46 \\ 
 \textbf{LL-FGSM} & \textbf{1} & \textbf{49.9} & \textbf{16.61} \\
 LL-I-FGSM & 1 & 46.5 & 10.86 \\ 
 LL-MI-FGSM & 1 & 48.4 & 14.06 \\ 
 LL-DI$^2$-FGSM & 1 & 48.8 & 14.70  \\
\rowcolor[rgb]{0.851,0.918,0.827} \textbf{Ours} & \textbf{1} & \textbf{50.1} & \textbf{17.07} \\ 
 \cmidrule{1-4}
\rowcolor[rgb]{0.851,0.918,0.827} \textbf{Ours} & \textbf{10} & \textbf{57.4}  \improvement{+7.5}  & \textbf{29.14}~ \improvement{+12.4}   \\ 
\cmidrule{1-4}
\multicolumn{4}{c}{\textbf{Query-based Methods}} \\
\cmidrule{1-4}
 \textbf{Targeted LL-FGSM} & \textbf{10} & 53.9 & 23.27 \\ 
\rowcolor[rgb]{0.851,0.918,0.827} \textbf{Ours} & \textbf{10} & \textbf{57.4}  \improvement{+3.5}  & \textbf{29.14}~ \improvement{+5.9}   \\ 
 \textbf{Targeted LL-FGSM} & \textbf{100\textsuperscript{\textdagger}} & 54.7 & 24.51 \\ 
\rowcolor[rgb]{0.851,0.918,0.827} \textbf{Ours} & \textbf{100} & \textbf{59.6}~\improvement{+5.2}  & \textbf{32.81}~ \improvement{+8.3}   \\ 
 \textbf{Random Perturbations} & \textbf{1,000} &  43.1	& 5.42  \\ 
\rowcolor[rgb]{0.851,0.918,0.827}\textbf{Ours~} & \textbf{1,000} & \textbf{60.8}~\improvement{+6.1}  & \textbf{34.73}~ \improvement{+10.2}   \\
\bottomrule
\end{tabular}
\caption{Transferring attacks from UCF$\rightarrow$Kinetics.  \\  \textsuperscript{\textdagger} - Maximum possible queries using Targeted LL-FGSM. \label{table:k400}}
\end{table}


\begin{table*}[t]
\renewcommand\thetable{3}
	\begin{subtable}[t]{.19\linewidth}
		\tablestyle{4pt}{1.1}
		{		\caption{\textbf{Target Architecture}} \label{tab:ablation-tgt}
			\begin{tabular}[t]{@{}l|cc@{}}
				Target Arch. & ASR & DR  \\
				\hline
				\textbf{R-(2+1)-D} & \textbf{58.7} & \textbf{31.3}  \\
				R3D       &  62.6 & 35.0 \\
				MC3       &  57.2 & 26.9  \\
				S3D       & 69.5 & 29.3  \\
				MViTv1    & 36.0 & 13.5  \\
				MViTv2    & 30.6 & 9.4  \\
			\end{tabular}
			
		}		
	\end{subtable}
	\begin{subtable}[t]{.19\linewidth}
		\tablestyle{4pt}{1.1}
		{		\caption{\textbf{Source Dataset}} \label{tab:ablation-src}
			\begin{tabular}[t]{@{}l|cc@{}}
				 Dataset & ASR & DR  \\
				\hline
				\textbf{UCF101} & \textbf{58.7} & \textbf{31.3} \\
				HMDB51          & 51.4 & 19.1 \\
			\end{tabular}
			
		}		
	\end{subtable}
		\begin{subtable}[t]{.19\linewidth}
		\tablestyle{4pt}{1.1}
		{		\caption{\textbf{Number of Queries}} \label{tab:ablation-query}
			\begin{tabular}[t]{@{}r|cc@{}}
				\# Queries  & ASR & DR  \\
				\hline
				1    &  50.1 & 17.1     \\
				10   &  56.1 & 26.8 \\
				\textbf{100}  &  \textbf{58.7} & \textbf{31.3} \\
				500  &  59.8 & 33.2 \\
				1000 &  60.8 & 34.8 \\
			\end{tabular}
			
		}		
	\end{subtable}
	\begin{subtable}[t]{.19\linewidth}\centering
		\tablestyle{4pt}{1.1}
		{		\caption{\textbf{Attacked Blocks}} \label{tab:ablation-layer}
			\begin{tabular}[t]{@{}r|cc@{}}
				Blocks & ASR & DR  \\
				\hline
				\textbf{4}     & \textbf{58.7} & \textbf{31.3}   \\
				3+4   & 59.1 & 31.9   \\
				2+3+4 & 58.8 & 31.4   \\
			\end{tabular}
			
		}		
	\end{subtable}
	\begin{subtable}[t]{.19\linewidth}
		\tablestyle{4pt}{1.1}
		{		\caption{\textbf{Attack Direction Search}} \label{tab:ablation-orth}
			\begin{tabular}[t]{@{}l|cc@{}}
				    & ASR & DR \\
				\hline
				\textbf{Random}     & \textbf{58.7} & \textbf{31.3} \\
				Orthogonal & 59.6 & 32.9 \\
			\end{tabular}
			
		}		
	\end{subtable}
	\captionsetup{width=.95\linewidth}
	\vspace{-2mm}
	\caption{\textbf{Ablation studies:} Attacking Kinetics400 classification models. {\em Default} setting is in \textbf{bold}. See Sec.~\ref{sec:ablation} for discussion.}\label{tab:ablation}
	\vspace{-2mm}
\end{table*}

\subsection{Targeting Kinetics400 Models}
\vspace{-1mm}

\begin{figure}[t]
    \centering
    \includegraphics[width=\linewidth]{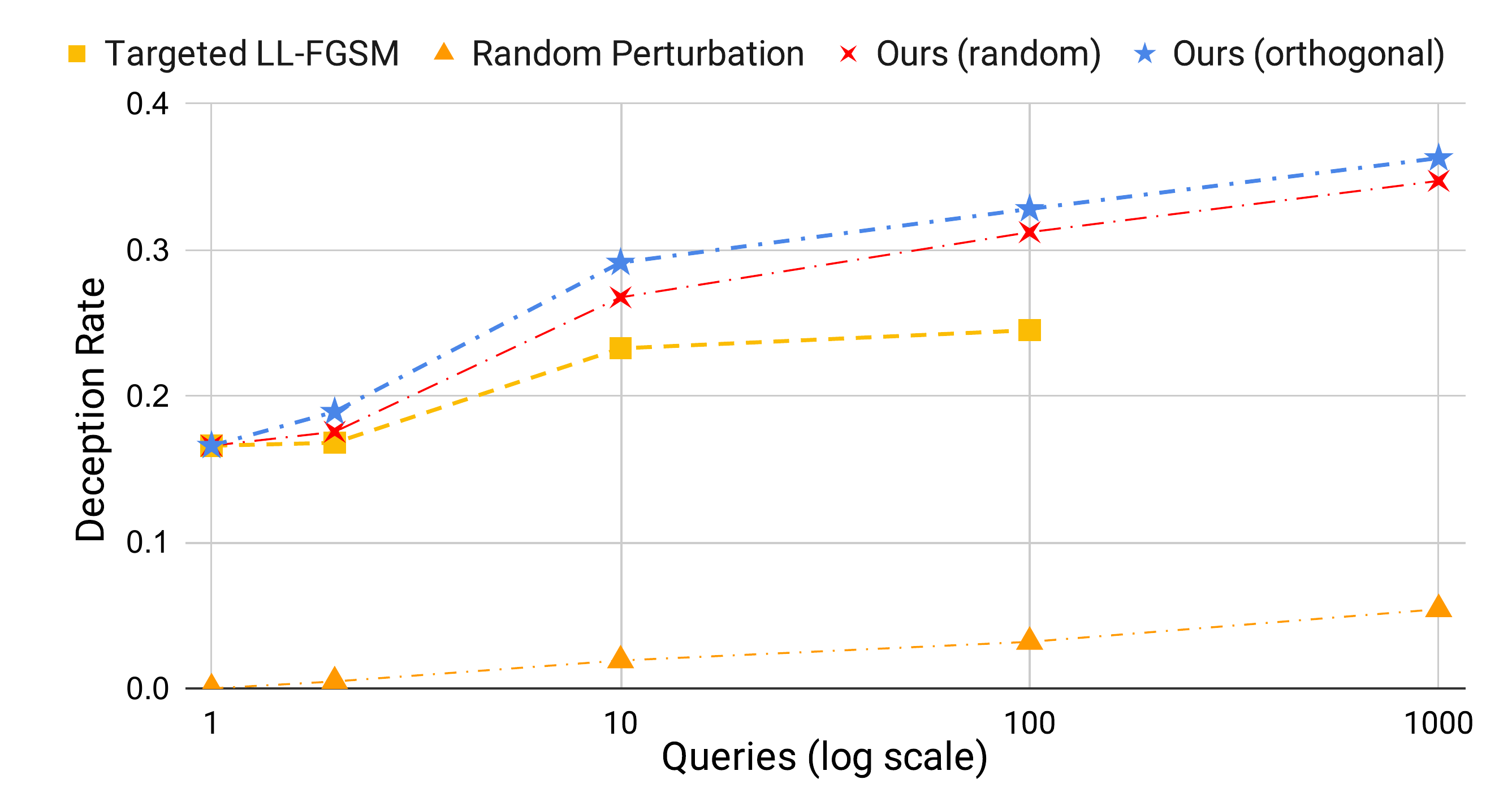}
    \vspace{-6mm}
    \caption{Video Representation Attack on Kinetics400 achieves higher deception rates than Targeted LL-FGSM and Random Perturbation baselines, especially in higher query settings. VRA with Orthogonal Attack Direction search outperforms random search.}
    \label{fig:queries}
\end{figure}

We evaluate our Video Representation Attack and all the other baselines in the UCF $\rightarrow$ Kinetics transfer setting with both source and target models being R-(2+1)-D 18 layer architecture. The results for attack magnitude $||\epsilon||_{\infty} = 4/255$ are presented in Table~\ref{table:k400}. We find that multi-iterated attacks do not transfer well across datasets even with the use of momentum boosting (MI-FGSM) and input augmentations (DI\textsuperscript{2}-FGSM)  to prevent overfitting. The single iteration Least Likely FGSM attack is the strongest transfer-based baseline. Our video representation attack is able to match this with a single query. However, as our attack can be extended to any number of queries, it is able to outperform the transfer-based baselines by 12.4\% in deception rate and 7.5\% in attack success rate using just 10 queries. In the multi-query setting, with 100 queries, our model outperforms the Targeted LL-FGSM by 5.2\% in ASR and 8.3\% in DR. Targeted LL-FGSM can only use 100 queries as it is limited by the number of classes in the training dataset (UCF101), while our VRA can be extended to 1,000 queries and achieve a deception rate of 34.7\%, which is 10\% higher than the baseline. See Fig.~\ref{fig:queries} for a comparison of deception rates in five different query settings: \{1, 2, 10, 100, 1000\}. 

\vspace{-1mm}
\subsection{Targeting Something-Somethingv2 Models}
\vspace{-1mm}

Something-Something primarily consists of actions done with objects, e.g. ``Putting [something] onto [something]". This distinguishes it from UCF and Kinetics where most of the actions are centered around people or groups of people. As a result domain shift between UCF and SSv2 is much higher and transferring attacks is much more challenging. 

For SSv2 we target a 50 layer R3D model using our R-(2+1)-D-18 model trained on UCF101. Our VRA attack (see Table~\ref{table:ssv2}) is able to outperform Targeted LL-FGSM by 2.9\% in ASR and 5.3\% in DR using 100 queries. Extended further to 1,000 queries it can achieve a deception rate 7.6\% higher than the best possible with Targeted LL-FGSM.

\begin{table}[h]
\renewcommand\thetable{2}
\tablestyle{2pt}{0.95}
\centering
\small
\begin{tabular}{lccc} 
\toprule
\textbf{Method} & \textbf{Queries} & \textbf{ASR}   & \textbf{DR}   \\ 
\toprule
\multicolumn{4}{c}{\textbf{Transfer-based Methods}} \\ 
\cmidrule{1-4}
Flickering\textbf{ }(\textbf{$\epsilon$ = 0.2})~\cite{Pony_2021_CVPR} & 1 & 46.7 & 1.09 \\ 
 FGSM & 1 & 46.7 & 1.09 \\
 \textbf{LL-FGSM} & \textbf{1} & 47.1 & 1.82 \\
\rowcolor[rgb]{0.851,0.918,0.827} \textbf{Ours} & \textbf{1} & \textbf{47.6} & \textbf{2.79} \\ 
 \cmidrule{1-4}
\rowcolor[rgb]{0.851,0.918,0.827} \textbf{Ours} & \textbf{10} & \textbf{51.0}  \improvement{+3.9}  & \textbf{9.09}~ \improvement{+7.3}   \\ 
\cmidrule{1-4}
\multicolumn{4}{c}{\textbf{Query-based Methods}} \\
\cmidrule{1-4}
 \textbf{Targeted LL-FGSM} & \textbf{10} & 49.8 & 6.73 \\ 
\rowcolor[rgb]{0.851,0.918,0.827} \textbf{Ours} & \textbf{10} & \textbf{51.0}  \improvement{+1.2}  & \textbf{9.09}~ \improvement{+2.4}   \\ 
 \textbf{Targeted LL-FGSM} & \textbf{100\textsuperscript{\textdagger}} & 50.4 & 8.00 \\ 
\rowcolor[rgb]{0.851,0.918,0.827} \textbf{Ours} & \textbf{100} & \textbf{53.3}  \improvement{+2.9}  & \textbf{13.26}~\improvement{+5.3}   \\ 
 \textbf{Random Perturbations} & \textbf{1,000} & 49.4 & 6.02 \\ 
\rowcolor[rgb]{0.851,0.918,0.827}\textbf{Ours~} & \textbf{1,000} & \textbf{54.6}~\improvement{+4.2}  & \textbf{15.64}~\improvement{+7.6}   \\
\bottomrule
\end{tabular}
\vspace{-1mm}
\caption{Transferring attacks from UCF $\rightarrow$ Smthng-Smthngv2. \\  \textsuperscript{\textdagger} - Maximum possible queries using Targeted LL-FGSM. \label{table:ssv2}}
\end{table}

\subsection{Ablation Studies}
\vspace{-2mm}
\label{sec:ablation}


We demonstrate the wide applicability of our method by (a) attacking six different pre-trained Kinetics classifiers, (b) initiating the attack using models trained on two different source datasets (UCF and HMDB), (c) demonstrating higher success rate with more queries, (d) generating the attack using multiple layers of the source model and (e) higher effectiveness of orthogonal attack direction search.

\vspace{2mm}
 \noindent \textbf{Target Model Architecture:} We find that while the attack is similarly effective across 4 different CNN architectures (R-(2+1)-D, R3D, MC3 and S3D), it is less effective against transformers (MViT v1 and v2). (See Table~\ref{tab:ablation-tgt}) This is expected as prior works~\cite{naseer2021intriguing, shao2022on, bai2021transformers, mahmood2021robustness} have reported that vision transformer architectures are more robust to adversarial perturbations than CNNs. 

 \noindent \textbf{Source Dataset:} Source models trained on UCF101 generate more effective attacks than models trained on the smaller HMDB51 dataset.  (See Table~\ref{tab:ablation-src})
 
  \noindent \textbf{Number of Queries:} While Higher number of queries are more effective (Table~\ref{tab:ablation-query}), we use 100 queries as a standard as it provides a good speed-effectiveness trade-off. 
  
  \noindent \textbf{Attacked Layers:} Our attack can be extended to attack multiple layers, attacking additional blocks of the source R-(2+1)-D model leads to marginally better results than simply attacking the final layer features. (Table~\ref{tab:ablation-layer})

  \noindent \textbf{Orthogonal Attack Direction Search} raises the deception rate by 1.6\% over random search. (Table~\ref{tab:ablation-orth})

\vspace{-2mm}
\section{Conclusion}
\label{sec:conclusion}
\vspace{-2mm}

We introduced a novel black-box attack which relies on similarity of learned representation between different video classification models. Our video representation attack achieved adversarial transferability across action recognition models trained on different datasets with limited domain overlap. Our attack combined the capabilities of transfer-based and query-based attacks to achieve higher attack success rates than prior baselines on large scale video-classification datasets in a challenging limited query hard-label feedback setting. 

\section*{Acknowledgements}

This work was supported in part by the Defense Advanced Research Projects Agency (DARPA) under Agreement HR00112090095. Dr. Naveed Akhtar is the recipient of Office of National Intelligence, National Intelligence Postdoctoral Grant (project number NIPG-2021-001) funded by the Australian Government. Professor Ajmal Mian is the recipient of an Australian Research Council Future Fellowship Award (project number FT210100268) funded by the Australian Government.

The authors would also like to thank Ishan Dave for discussions on action recognition learning methods.


{\small
\bibliographystyle{ieee_fullname}
\bibliography{arxiv}
}

\appendix

\section{Implementation Details}
\label{sec:implement}

\noindent{\textbf{Training of Source Models}}: We train source models for classifying UCF101 and HMDB51 classes. For most experiments, the R-(2+1)-D architecture with 18 layers is used, however for ablation studies we also experiment with other architectures (See Table 2 of the main paper). A 100 epoch single cycle cosine annealing learning rate scheduler is used, with a peak learning rate of 0.01. An initial warmup period of 5 cycles is also employed, in accordance with best practices from the literature. We use a batch size of 32 videos, with 16 frames per video clip. Random cropping and Flipping are the only data augmentations used during training. 


\section{Datasets}
\label{sec:assets}

\begin{figure}[h]
\centering
\includegraphics[width=0.8\linewidth]{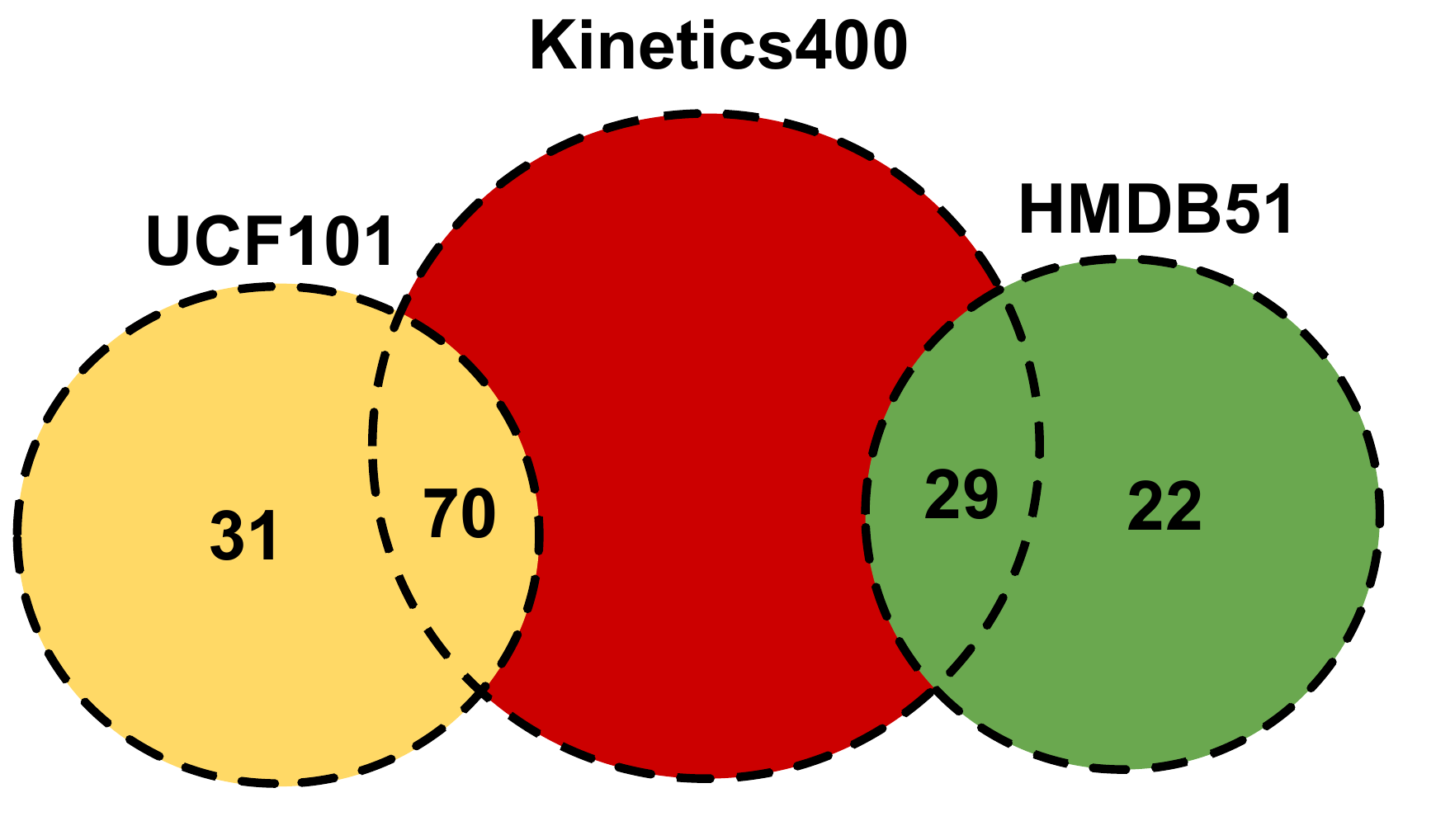}
\caption{Overlap between the datasets used. There is a small amount of overlap between classes in HMDB and UCF, but that is ignored for this visualization.}
\label{fig1}
\end{figure}

We utilize UCF101 and HMDB51 datasets for training our source models, whereas the test set for Kinetics400 is used for testing the attacks on black box models (see Figure \ref{fig1}). 

\noindent\textbf{UCF101}~\cite{soomro2012ucf101} has around 13,320 web videos representing 101 different human activities such as Sports, etc.

\noindent\textbf{HMDB51}~\cite{Kuehne2011hmdb} is a somewhat smaller dataset with 6,849 total videos belonging to 51 different human action categories.

\noindent\textbf{Kinetics400}~\cite{kay2017kinetics} is a much bigger dataset, however it is distributed by the original creators only in the form of a metadata file with video URLs and annotations, and needs to be crawled by the end user, as a result the number of videos may vary. Our crawled version of the test set has about 18,000 videos covering all 400 classes, and this is the dataset used for all experiments in this work.

\noindent\textbf{Something-Something-V2}~\cite{goyal2017something} is distinct from the other datasets because it primarily focuses on object centric actions, e.g. "Pushing something from left to right", "Turning something upside down", etc.

Each of these datasets can be obtained from the following sources:

\textbf{UCF101}: \url{https://www.crcv.ucf.edu/data/UCF101.php}

\textbf{HMDB51}: \url{https://serre-lab.clps.brown.edu/resource/hmdb-a-large-human-motion-database/}

\textbf{Kinetics400 Labels}: \url{https://deepmind.com/research/open-source/kinetics}

\textbf{Kinetics400 Crawler}: \url{https://github.com/activitynet/ActivityNet/tree/master/Crawler/Kinetics}

\textbf{Something-Something-V2}: \url{https://developer.qualcomm.com/software/ai-datasets/something-something}

\section{Pre-Trained Models}
\label{sec:models}

\noindent{\textbf{Pre-Trained Black box Models}} are obtained from the TorchVision and PyTorchVideo model zoos, and the appropriate pre-processing operations are applied to videos before querying the models. Single clip clean accuracues for the models used are listed in Table~\ref{tab:models}.

\textbf{TorchVision Model Zoo (Video Classification)}: \url{https://pytorch.org/vision/stable/models.html#video-classification}

\textbf{PyTorch Video Model Zoo}: \url{https://pytorchvideo.readthedocs.io/en/latest/model_zoo.html}


\begin{table}[h]
\centering
\setlength\tabcolsep{2pt}
\begin{tabular}{lccc}
\hline

\hline

\hline\\[-3mm]
\textbf{Model} & \textbf{\# Frames} & \textbf{Resolution} &\textbf{ Top-1 Acc.} \\
\midrule
\multicolumn{4}{l}{\textbf{Something-Something-v2}} \\
R3D-50       & 8  & 224 & 53.869  \\
\midrule
\textbf{Kinetics-400} & & \\
R-(2+1)-D-18 & 16 & 112 & 60.115  \\
R3D-18       & 16 & 112 & 57.526  \\
MC3-18       & 16 & 112 & 58.485  \\
S3D-18       & 42~\textsuperscript{$\dagger$}& 224 & 43.145  \\
S3D-18       & 128 & 224 & 49.952 \\
MViTv1-B     & 16 & 224 & 73.538  \\
MViTv2-S     & 16 & 224 & 76.606  \\
\bottomrule
\end{tabular}
\caption{Clean Accuracy of pretrained black-box targeet models.}
\label{tab:models}
\end{table}

\section{Controlled Class Overlap Experiment}
\label{sec:control}

In the main paper we present results for attacking black-box Kinetics400~\cite{kay2017kinetics} Action Recognition Models using source models trained on the full UCF101~\cite{soomro2012ucf101} and HMDB51~\cite{Kuehne2011hmdb} datasets. UCF has 70 classes in common with Kinetics, whereas HMDB has 29. However, the extent of this overlap is fixed and hence does not allow us to study the impact of class overlap on transferability of adversarial perturbations. In this section, we devise a new set of experiments where we train source models on a subset of UCF and HMDB datasets with a controlled amount of overlap with Kinetics classes and study the success of attacks generated using these source models.

We carry out the experiments using the following set of source models datasets:

\begin{enumerate}
\vspace{-2mm}
    \item {\bf UCF101:} The complete UCF101 dataset (70\% class overlap with Kinetics).
\vspace{-2mm}
    \item {\bf UCF70c:} Only the 70 classes common between UCF and Kinetics (100\% class overlap with Kinetics).
    \vspace{-2mm}
\item {\bf UCF31nc:} Only the 31 classes present in UCF but not in Kinetics (0 class overlap with Kinetics).
\vspace{-2mm}
    \item {\bf UCF41m:} UCF31nc + 10 common classes (10 class overlap with Kinetics).
\vspace{-2mm}
    \item {\bf UCF61m:} UCF31nc + 30 common classes (30 class overlap with Kinetics).
\vspace{-2mm}
    \item {\bf UCF81m:} UCF31nc + 50 common classes (50 class overlap with Kinetics).
\vspace{-2mm}
    \item {\bf HMDB51:} The complete HMDB51 dataset (57\% class overlap with Kinetics)
\vspace{-2mm}
    \item {\bf HMDB22nc:} Only the 22 classes present in HMDB but not in Kinetics(0 class overlap  with Kinetics)
\vspace{-2mm}
    \item \textbf{HMDB32m:} HMDB22nc + 10 common classes (10 class overlap with Kinetics).
\end{enumerate}


For this attack setting, we utilize 100 queries of the black-box per attack and perturbations of $L_\infty$ norm of $4/255$. Results for these attacks can be seen in Table~\ref{tab:resultsoverlap}. There is a clear correlation between extent of overlap and the deception rate. However, even in case where the overlap between source and target dataset is zero, the attack does achieve a fooling rate that is better than a simple random noise baseline.

\begin{table}[h]
\small
\centering
\setlength\tabcolsep{1pt}
\begingroup
\renewcommand{\arraystretch}{1.0}
\begin{tabular}{cccc}
\hline

\hline

\hline\\[-3mm]
\textbf{Attack Setting}  & \textbf{\# source} & \textbf{\# common} & \textbf{Deception}  \\
  & \textbf{classes} & \textbf{classes} & \textbf{Rate}  \\
\midrule
Random Noise Baseline & & & 2\% \\ 
\midrule
UCF101 $\rightarrow$ Kinetics &101 &70 & 32\%  \\
UCF70c $\rightarrow$ Kinetics  &70 &70 & 30\%  \\
UCF31nc $\rightarrow$ Kinetics &31 &0 & 16\%  \\
UCF41m $\rightarrow$ Kinetics  &41 &10 & 22\%  \\
UCF61m $\rightarrow$ Kinetics  &61 &30 & 25\%  \\
UCF81m $\rightarrow$ Kinetics  &81 &50 & 30\%  \\
\midrule
HMDB51 $\rightarrow$ Kinetics &51 &29 & 19\%  \\
HMDB29c $\rightarrow$ Kinetics &29 &29 & 15\%  \\
HMDB22nc $\rightarrow$ Kinetics &22 &0 & 10\% \\
HMDB32m $\rightarrow$ Kinetics  &32 &10 & 13\%  \\

\bottomrule
\end{tabular}
\endgroup
\caption{Using source models trained with different subsets of UCF101 and HMDB51 to attack a black box R-(2+1)-D Kinetics400 model. A maximum of 100 queries are used per attack and perturbations are restricted to a $L_\infty$ norm of $4/255$.}
\label{tab:resultsoverlap}
 \vspace{3mm}
\end{table}

\section{Perturbations with Additional Constraints}
\label{sec:constraints}

As discussed in Section 3.4 of the main paper, we can impose additional constraints such as sparsity~\cite{wei2019sparse} and temporal smoothness~\cite{Pony_2021_CVPR} on the perturbations generated by our Video Representation Attack. In this section, we present results for applying the sparsity constraint to the adversarial perturbations. Since the sparsity loss depends on the perturbation itself, computing the perturbation becomes an iterated process. We use a fixed number of iterations $n$ for this process, and following prior iterated attacks utilize a modified perturbation magnitude $\alpha = \epsilon/n$ in the generation process. In order to ensure spatio-temporal sparsity, simple $L_1$ norm is used as the sparsity loss. The sparsity term and the representation attack term weights are adjusted using the sparsity parameter $\lambda$ .

\begin{equation}
\begin{aligned}
    \Delta x_{i+1} = - \alpha \cdot sign(\nabla_x (\mathcal{L}_{VRA}(f_s^+(x + \Delta x_{i}), \vec{\mathbf{v}})) \\
    + \lambda \cdot \mathcal{L}_{sparsity}(\Delta x_{i})).
    \label{eq:pertrand}
\end{aligned}
\end{equation}

Results of our  
experiments 
are presented in Table~\ref{tab:resultssparse}. Adding the sparsity objective results in perturbations that are sparser (as measured by their $L_1$ norm), but less transferable. 

\begin{table}[h]
\small
\centering
\setlength\tabcolsep{1pt}
\begin{tabular}{lcc}
\hline

\hline

\hline\\[-3mm]
\textbf{Attack Setting}  & \textbf{Deception Rate}  & $||\Delta x||_1$ \\
\midrule
UCF101 $\rightarrow$ Kinetics ($\lambda = 0$)  & 32\% & 4.2 \\
UCF101 $\rightarrow$ Kinetics ($\lambda = 0.0001$)  & 27\% & 3.9 \\
UCF101 $\rightarrow$ Kinetics ($\lambda = 0.001$)  & 25\%  & 3.4\\
\bottomrule
\end{tabular}
\vspace{-2mm}
\caption{Attacking a black box R-(2+1)-D Kinetics400 action recognition model, with \textbf{sparse} perturbations restricted to a $L_\infty$ norm of $4/255$.}
\label{tab:resultssparse}
 \vspace{3mm}
\end{table}

\section{Effect of Self-Supervised Pre-Trained Source Model}
\label{sec:selfsup}

Recent work on constrastive self-supervised pre-training of action recognition models such as CVRL~\cite{qian2021spatiotemporal} and TCLR~\cite{dave2022tclr} demonstrates that significant improvements in classification accuracy. Additionally, since these models initially learn to extract features from videos without the guidance of class labels, we expect that such source models will result in qualitatively different transferability characteristics for adversarial perturbations.

We utilize TCLR~\cite{dave2022tclr} pre-training on UCF101 for our experiments. As previously, we limit our experiument to 5 queries of the black-box per attack (i.e. $Q_{max} = 4$) and perturbations of $L_\infty$ norm of $4/255$. 

As can be seen from Table~\ref{tab:resultsssl} the pre-trained models are able to generate stronger attacks. We hypothesize that the reason for this strength is that spatio-temporal features learned by models trained using self-supervised learning are not tied to the label space of the source dataset, but are more general.

\begin{table}[h]
\small
\centering
\setlength\tabcolsep{1pt}
\begin{tabular}{lc}
\hline

\hline

\hline\\[-3mm]
\textbf{Attack Setting}  & \textbf{Deception Rate}  \\
\midrule
UCF101 $\rightarrow$ Kinetics  & 32\%  \\
UCF101 (TCLR Pre-Trained) $\rightarrow$ Kinetics  & 36\%  \\
\midrule
HMDB51 $\rightarrow$ Kinetics  & 19\%  \\
HMDB51 (TCLR Pre-Trained) $\rightarrow$ Kinetics  & 31\%  \\

\bottomrule
\end{tabular}
\vspace{-2mm}
\caption{Using source models pretrained using TCLR~\cite{dave2022tclr} self-supervised pretraining to attack a black box R-(2+1)-D Kinetics400 action recognition model, with perturbations restricted to a $L_\infty$ norm of $4/255$.}
\label{tab:resultsssl}
 \vspace{3mm}
\end{table}

\section{Attack Visualizations}
\label{sec:viz}

\begin{figure*}[t]
\begin{subfigure}[b]{0.45\linewidth}
    \centering
    \includegraphics[width=0.3\linewidth]{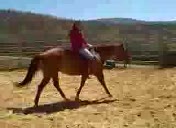}
    \includegraphics[width=0.3\linewidth]{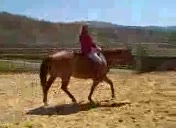}
    \includegraphics[width=0.3\linewidth]{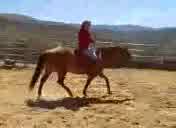} \\
    \includegraphics[width=0.3\linewidth]{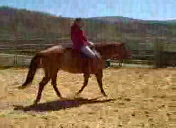}
    \includegraphics[width=0.3\linewidth]{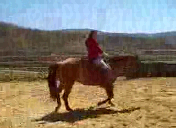}
    \includegraphics[width=0.3\linewidth]{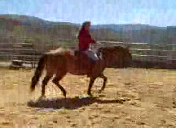} \\ 
    \includegraphics[width=0.3\linewidth]{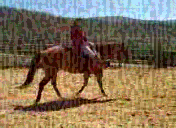}
    \includegraphics[width=0.3\linewidth]{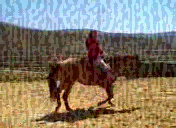}
    \includegraphics[width=0.3\linewidth]{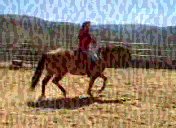}
    \caption{\texttt{HorseRiding}}
\end{subfigure}
\hfill
\begin{subfigure}[b]{0.45\linewidth}
    \centering
    \includegraphics[width=0.3\linewidth]{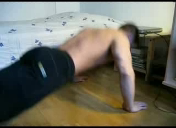}
    \includegraphics[width=0.3\linewidth]{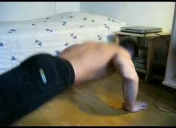}
    \includegraphics[width=0.3\linewidth]{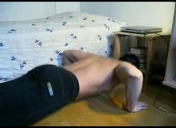} \\
    \includegraphics[width=0.3\linewidth]{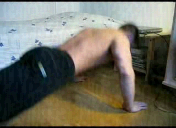}
    \includegraphics[width=0.3\linewidth]{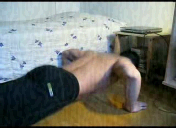}
    \includegraphics[width=0.3\linewidth]{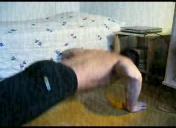} \\ 
    \includegraphics[width=0.3\linewidth]{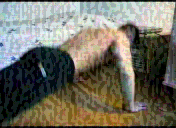}
    \includegraphics[width=0.3\linewidth]{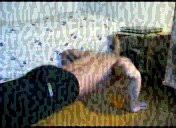}
    \includegraphics[width=0.3\linewidth]{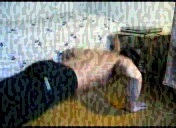}
    \caption{\texttt{PushUps}}
\end{subfigure}
\caption{Visualization of perturbed videos. \textbf{Top Row:} Clean video frames, \textbf{Middle Row:} Perturbed frames ($\epsilon = 4$), \textbf{Bottom Row:} Perturbed frames ($\epsilon = 16$). GIFs included in folder GIFs. (Best viewed in color.)}
\label{fig:viz}
\end{figure*}

We provide some samples of perturbed video frames generated using the proposed attack in Figure~\ref{fig:viz}. As can be observed from these video frames, attacks of perturbation magnitudes upto $L_\infty = 4$ are nearly imperceptible to the human eye, whereas of $L_\infty = 16$ can be clearly observed.

In order to examine the impact of motion on the perceptibility of the perturbations, GIFs of perturbed videos are also included with this supplementary material. Five different samples are included at locations \texttt{/GIFs/<ClassName>}. Clean videos and Perturbed Videos of perturbation magnitudes  $L_\infty \in \{ 4, 16 \}$ are included.

\end{document}